\documentclass[letterpaper]{article} 
\usepackage{aaai23}  
\usepackage{times}  
\usepackage{helvet}  
\usepackage{courier}  
\usepackage[hyphens]{url}  
\usepackage{graphicx} 
\usepackage{natbib}  
\usepackage{caption} 
\usepackage{algorithm}
\usepackage{algorithmic}

\usepackage{newfloat}
\usepackage{listings}
\DeclareCaptionStyle{ruled}{labelfont=normalfont,labelsep=colon,strut=off} 
\floatstyle{ruled}
\newfloat{listing}{tb}{lst}{}
\floatname{listing}{Listing}
\title{CasFusionNet: A Cascaded Network for Point Cloud Semantic Scene Completion by Dense Feature Fusion}
\author{
Jinfeng Xu, Xianzhi Li\thanks{Corresponding author}, Yuan Tang, Qiao Yu, Yixue Hao, Long Hu, Min Chen
}
\affiliations{
Huazhong University of Science and Technology\\

jinfengxu.edu@gmail.com,\{xzli, yuan\_tang, qiaoyu\_epic, yixuehao, hulong, minchen2012\}@hust.edu.cn
}

\usepackage{bibentry}
\usepackage{amsfonts} 
\usepackage{booktabs}
\usepackage{multirow}

\begin{document}

\maketitle

\begin{abstract}
	
	Semantic scene completion (SSC) aims to complete a partial 3D scene and predict its semantics simultaneously. 
	Most existing works adopt the voxel representations, thus suffering from the growth of memory and computation cost as the voxel resolution increases.
	Though a few works attempt to solve SSC from the perspective of 3D point clouds, they have not fully exploited the correlation and complementarity between the two tasks of scene completion and semantic segmentation.
	In our work, we present CasFusionNet, a novel cascaded network for point cloud semantic scene completion by dense feature fusion.
	Specifically, we design (i) a global completion module (GCM) to produce an upsampled and completed but coarse point set, (ii) a semantic segmentation module (SSM) to predict the per-point semantic labels of the completed points generated by GCM, and (iii) a local refinement module (LRM) to further refine the coarse completed points and the associated labels from a local perspective.
	We organize the above three modules via dense feature fusion in each level, and cascade a total of four levels, where we also employ feature fusion between each level for sufficient information usage.
	Both quantitative and qualitative results on our compiled two point-based datasets validate the effectiveness and superiority of our CasFusionNet compared to state-of-the-art methods in terms of both scene completion and semantic segmentation.
	The codes and datasets are available at: https://github.com/JinfengX/CasFusionNet.
	
\end{abstract}

\section{Introduction}

\label{sec:intro}
\begin{figure}[t]
	\centering
	\includegraphics[width=\linewidth,scale=1.00]{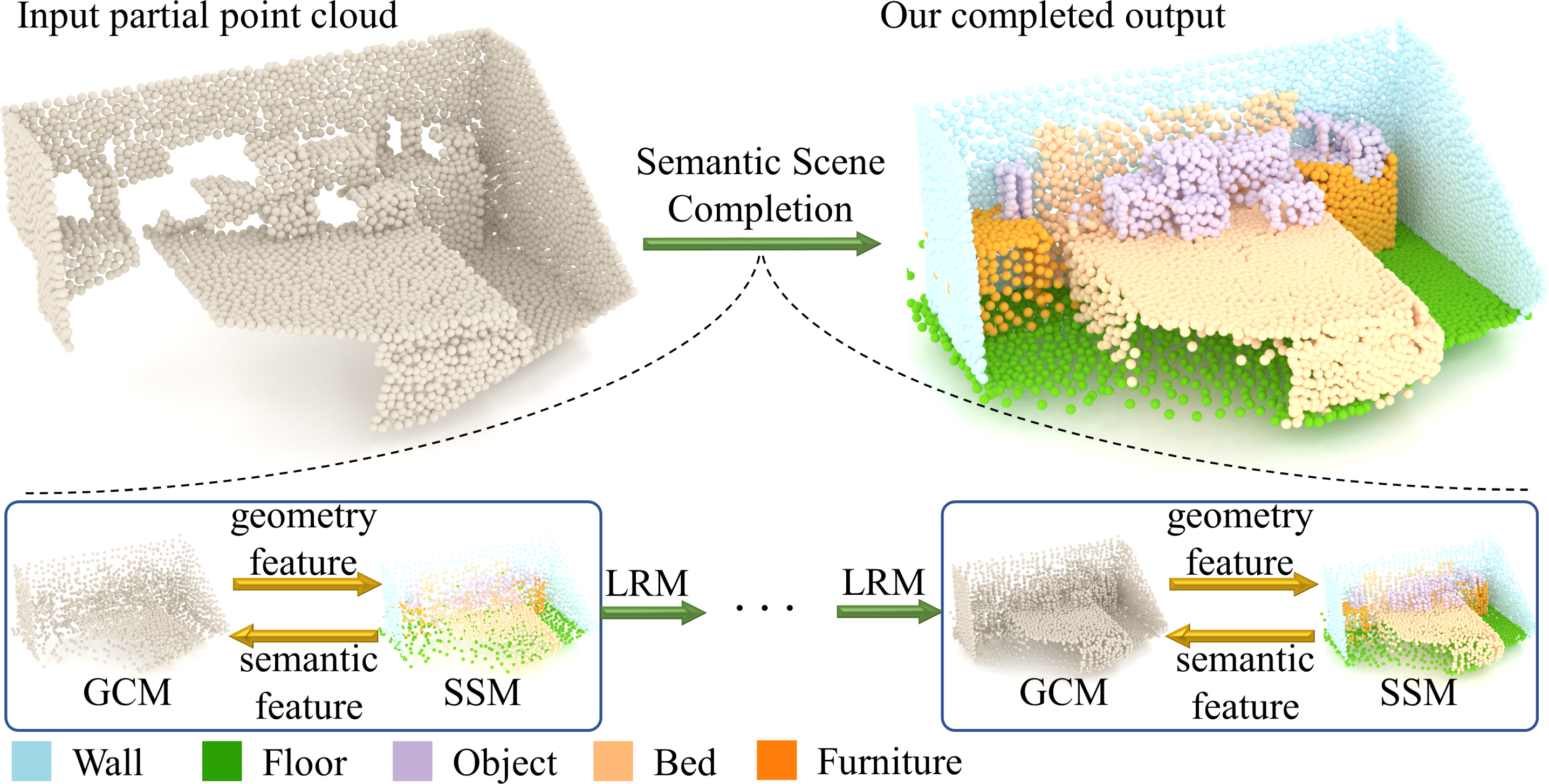}
	\caption{Given a partial point-based scene (left), our CasFusionNet completes and segments the scene (right) by cascaded levels. Each level is composed of GCM, SSM and LRM, which are connected by dense feature fusion.}
	\label{fig:teaser}
\end{figure}

Humans can infer the complete shapes and semantics from only a partial observation of a 3D scene based on experience. 
To enable an intelligent agent to behave like humans in 3D physical world, semantic scene completion (SSC) is gaining more attention with the goal of simultaneously reconstructing complete 3D scenes and predicting associate semantics from RGB-D images of a given viewpoint.
As SSC supplies complete 3D information of scenes, it benefits diverse applications, e.g., robot navigation and grasping~\cite{gupta2017cognitive,varley2017shape,liang2021sscnav}, automatic driving, high-quality visualization, etc.

Followed by the pioneer work SSCNet~\cite{song2017semantic}, most of existing methods~\cite{guo2018view,li2020attention,li2021anisotropic,cai2021semantic} utilize 3D CNN to predict the volumetric occupancy and semantics based on voxel representation.
Yet, these voxel-based networks suffer from tremendous computation and memory cost when the voxel resolution is high.
In contrast, point clouds, or scattered collections of points in 3D, are arguably the simplest shape representation, especially for large-scale 3D scenes with complex structures and fine details.
Thus, recent few works~\cite{zhang2021point,wang2022learning} attempt to consume only points as network input to handle SSC, where two branches are commonly designed, one for point-based scene completion and the other for point-based semantic segmentation. 
Despite notable achievements, existing works tend to ignore the high correlation and complementarity between scene completion and semantic segmentation, so there is still a lot of room for performance improvement.

In our work, we present CasFusionNet, a novel cascaded network for point-based SSC that associates scene completion with semantic segmentation by dense feature fusion. 
Figure~\ref{fig:teaser} shows the key idea.
Specifically, we design (i) a global completion module (GCM) to produce an upsampled and completed but maybe coarse point set from a partial 3D scene represented by points, (ii) a semantic segmentation module (SSM) to predict per-point semantic labels of the completed points generated by GCM, and (iii) a local refinement module (LRM) to further refine the coarse completed points and associated labels from a local perspective.
We organize the three modules via dense feature fusion in each level, and cascade a total of four levels, where we also employ skip connection and feature fusion between each level for sufficient information usage.

In our network, both the geometric features extracted from 3D points and the semantic features extracted from semantic label distribution are running through all the three designed modules in each level, which closely ties the tasks of scene completion and segmentation. 
To validate the effectiveness of our CasFusionNet, we prepare two point-based datasets based on existing SSC datasets, i.e. NYUCAD~\cite{firman2016structured} and  PCSSC-Net~\cite{zhang2021point}.
Extensive quantitative and qualitative results show the superiority and effectiveness of our network compared to state-of-the-arts; see Figure~\ref{fig:teaser} for an example output of our method.
Overall, our contributions are summarized as follows:
\begin{itemize}
	\item We propose a novel cascaded network (CasFusionNet) for point cloud semantic scene completion. It completes the partial scene and predicts the semantics simultaneously in a progressive manner by dense feature fusion.
	\item We design three novel modules, i.e. the global completion module (GCM), the semantic segmentation module (SSM) and the local refinement module (LRM), to complete, segment, and locally refine the scene, respectively.
	\item We contribute two point-based datasets for SSC task, and extensive experiments validate that our network outperforms previous works significantly.
\end{itemize}

\section{Related Work}
\label{sec:rw}

\newcommand{\para}[1]{\vspace{.05in}\noindent\textbf{#1}}

\para{Semantic Scene Completion.} 
With recent advances in 3D deep learning, semantic scene completion (SSC) has been widely explored, where both semantics and geometry are jointly inferred from a partial 3D scene.
To directly adopt 3D CNNs, most recent works encode a 3D scene as a 3D grid, in which cells describe semantic occupancy of the space.
For example, 
SSCNet~\cite{song2017semantic} was first proposed to tackle SSC with an end-to-end 3D convolution network, which predicts volumetric occupancy and semantic labels of scenes. 
Later, some followers~\cite{guo2018view, liu2018see, garbade2019two, li2020attention} leveraged the 2D semantic priors of color images via feature projection to improve the performance. 
Recent works~\cite{dourado2021edgenet, dourado2022data, wang2022ffnet} further fused complex features that are extracted from depth or color images to 2D semantic network. 
In addition, IMENet~\cite{li2021imenet} proposed an iterative fusion scheme to ensure the branches of 2D segmentation and 3D scene completion fully benefit each other. 
However, the cubic growth of computational and memory requirements of 3D CNN blocks the depth of networks, which limits the task performance.
To relieve the computational cost, various efficient network designs were introduced, such as spatial group convolution network~\cite{zhang2018efficient}, octree-based network~\cite{wang2020deep}, lightweight consecutive dimensional convolutions~\cite{li2019rgbd, li2020anisotropic, li2021anisotropic}, and efficient depth information embedding~\cite{chen20203d}.

Opposed to occupancy grids (voxels), 3D point cloud is a convenient and memory-efficient representation, which expresses geometry with fine details.
Only a few works have in fact explored point-based SSC.
Early method SPCNet~\cite{zhong2020semantic} applied point encoder-decoder architecture on ``pointlized" voxels.
To use both voxels and points, SISNet~\cite{cai2021semantic} converted voxels into points to recover detailed 3D shapes in scene-to-instance completion stage. 
A recent point-voxel aggregation strategy~\cite{tang2022not} used point-based network as the mainstream to improve the learning efficiency and capability of the framework. 
Yet, these methods are still confined by the low resolution and high operating cost of voxel representation. 

Very recently, PCSSC-Net~\cite{zhang2021point} tackled SSC solely based on point clouds and directly concatenated predicted semantic labels with the point features for better completion. 
Another method~\cite{wang2022learning} obtained semantic labels by attaching an extra point cloud segmentation network.
However, the two existing point-based methods do not fully exploit the connection between point completion and segmentation.
In our work, we argue that point completion and segmentation are complementary and highly correlated, thus motivating us to design CasFusionNet to encourage the features in completion and segmentation modules to fully communicate and merge.

\begin{figure*}[t]
	\centering
	\includegraphics[width=1\textwidth]{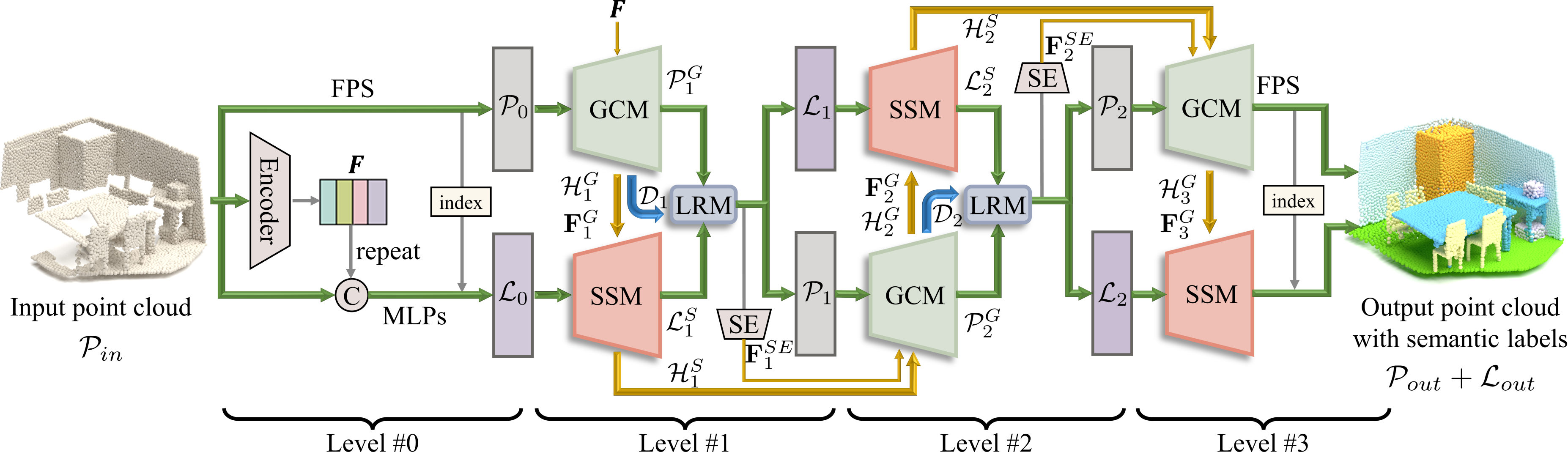}
	\caption{Illustrating the architecture of our CasFusionNet. Given a partial point cloud $\mathcal{P}_{in}$, our network semantically completes it via four successive levels in a coarse-to-fine manner and outputs a complete scene $\mathcal{P}_{out}$ with the associated per-point semantic labels $\mathcal{L}_{out}$. Each level is composed of a global completion module (GCM), a semantic segmentation module (SSM) and a local refinement module (LRM) to complete, segment, and refine the scene, which are closely connected by dense feature fusion.
	}
	\label{fig:architecture}
\end{figure*}

\para{Point Cloud Completion on Single Object.}
Though our work focuses on semantic scene completion, we here still briefly summarize the related works on single point cloud completion.
PCN~\cite{yuan2018pcn} was the pioneering work to address shape completion by folding a small patch of 2D grids for each point to represent the local geometry. Some followers~\cite{liu2020morphing,wen2020point,zong2021ashf} further improved the performance based on a similar folding-based strategy.
On the other hand,
coarse-to-fine methods~\cite{xie2020grnet,deng20213d, pan2021variational,zhang2021view} completed objects in an explicit and controllable way, which gradually increase density or change distribution of the point set. 
For instance, RFNet~\cite{huang2021rfnet}, PMP-Net~\cite{wen2021pmp} and PMP-Net++~\cite{wen2022pmp} completed the points level by level, where the recurrent neural network was utilized to reserve useful information of previous level.
ASFM-Net~\cite{xia2021asfm} and SnowflakeNet~\cite{xiang2022snowflake}  upsampled and moved the points at each refinement iteration.
Recently, a two-path network~\cite{zhao2021relationship} for pairwise completion was proposed to separately complete objects which bears a strong spatial relation.

Nevertheless, the above completion networks focus on small or single objects, which can hardly directly handle a large-scale incomplete 3D scene with occlusion and multiple kinds of objects.
Moreover, these single point cloud completion methods have no function of semantic segmentation.

\section{Method}
\label{sec:method}

\subsection{Overview of Network Architecture}
Given a partial point cloud $\mathcal{P}_{in} \in \mathbb{R}^{N\times 3}$ with $N$ points as input, which represents the incomplete 3D scene, our target is to attain a complete 3D scene $\mathcal{P}_{out} \in \mathbb{R}^{M\times 3}$ with $M$ points ($M$$\geq$$N$), as well as the predicted per-point semantic labels $\mathcal{L}_{out} \in \mathbb{R}^{M\times C}$, where $C$ is the total number of semantic classes.
Figure~\ref{fig:architecture} shows the overall pipeline of CasFusionNet, which semantically completes a scene via four successive levels in a coarse-to-fine manner.
The outputs of previous level are the inputs of next level; see the green arrows for cascading the four levels.
Further, the scene completion and semantic segmentation tasks are closely communicated and complement each other by dense feature fusion, both within and between levels; see the yellow arrows for features flow. 
Below, we shall elaborate on the details of each level.

In level \#0, we first feed $\mathcal{P}_{in}$ into an encoder to extract the hierarchical scene-wise features $\mathbf{F}$.
Empirically, the encoder is designed by using the set abstraction layer~\cite{qi2017pointnet++} and the point transformer layer~\cite{zhao2021point}.
Please refer to our supplementary material for the detailed architecture of the encoder.
To realize semantic completion, a naive following step is to decode $\mathbf{F}$ into a complete scene and its associated semantic labels.
However, the one-step operation is often difficult to yield high-quality and fine-grained predictions.
Hence, we propose to adopt the coarse-to-fine strategy by repeating semantic completion multiple times, thus allowing network to refine its predictions steadily.
To ensure enough repeating times (or levels), the computation consumption in level \#0 should not be too large, and it should not cause a huge calculation in the subsequent levels.
To this end, in the initial level, instead of directly upsampling and completing $\mathcal{P}_{in}$, we downsample it using the farthest point sampling (FPS) to obtain a sparse point cloud $\mathcal{P}_0$, which will be completed by subsequent levels.
The advantage of this operation is that, $\mathcal{P}_0$ has fewer points than $\mathcal{P}_{in}$, thus reducing the computation of following levels, but still preserving necessary geometrical structures.
In this level, we also predict the coarse per-point semantic labels $\mathcal{L}_0$ associated with $\mathcal{P}_0$ by using the duplicated $\mathbf{F}$ via multi-layer perceptions (MLPs); see Figure~\ref{fig:architecture} for details.

In level \#1, given $\mathcal{P}_0$ and $\mathcal{L}_0$, our purpose is to obtain the refined completed scene $\mathcal{P}_1$ and its semantic prediction $\mathcal{L}_1$.
As shown in Figure~\ref{fig:architecture}, this level is mainly composed of three modules, i.e. Global Completion Module (GCM), Semantic Segmentation Module (SSM), and Local Refinement Module (LRM).
Specifically, GCM consumes the point cloud produced by previous level as input and generates a coarse but completed scene by point displacements and upsampling (e.g., $\mathcal{P}_0 \rightarrow \mathcal{P}_1^G$ in level \#1).
Here, the upsampling ratio is $\lambda$ and set as 2 by default.
SSM aims to predict semantic labels of the completed scene (i.e. $\mathcal{P}_1^G$), which takes the hidden layer output $\mathcal{H}_1^G$ and the feature vector $\mathbf{F}_1^G$ produced by GCM, as well as the $\mathcal{L}_0$ as inputs.
We shall explain $\mathcal{H}_1^G$ and $\mathbf{F}_1^G$ in the following subsection.
To better recover a 3D scene with fine-grained geometric details, we further design LRM (to be detailed later), which conducts refinement locally according to the point displacement vectors $\mathcal{D}_1$ generated by GCM, and finally outputs $\mathcal{P}_1$ and $\mathcal{L}_1$.

The level \#2 continues to upsample and complete $\mathcal{P}_1$ into a finer result $\mathcal{P}_2$ and its associated labels $\mathcal{L}_2$.
Most operations are the same as in level \#1, except that we further feed the hidden layer output $\mathcal{H}_1^S$ of SSM and the scene-wise features $\mathbf{F}_1^{SE}$ extracted by shared encoder (SE) in the first level to the second level's GCM as well.
Intuitively, $\mathcal{H}_1^S$ encodes the semantic features extracted in the first level, while $\mathbf{F}_1^{SE}$ encodes the scene-wise global features embedded from the refined completed point cloud produced by the first level's LRM.
In this way, we can fully utilize the information of the first level.
Note that, we implement SE by using the same structure as the encoder in level \#0. 

Further, level \#3 repeats the above process again.
As shown in the right part of Figure~\ref{fig:architecture}, since the completion and segmentation results after the third level's GCM and SSM are good enough, we thus remove LRM in this level and use FPS to keep $M$ points as the final outputs.
Note that, FPS operation is optional, which can be used only when we have a specific requirement on the output point number.

\subsection{Global Completion Module}
\begin{figure}[t]
	\centering
	\includegraphics[width=\linewidth,scale=1.00]{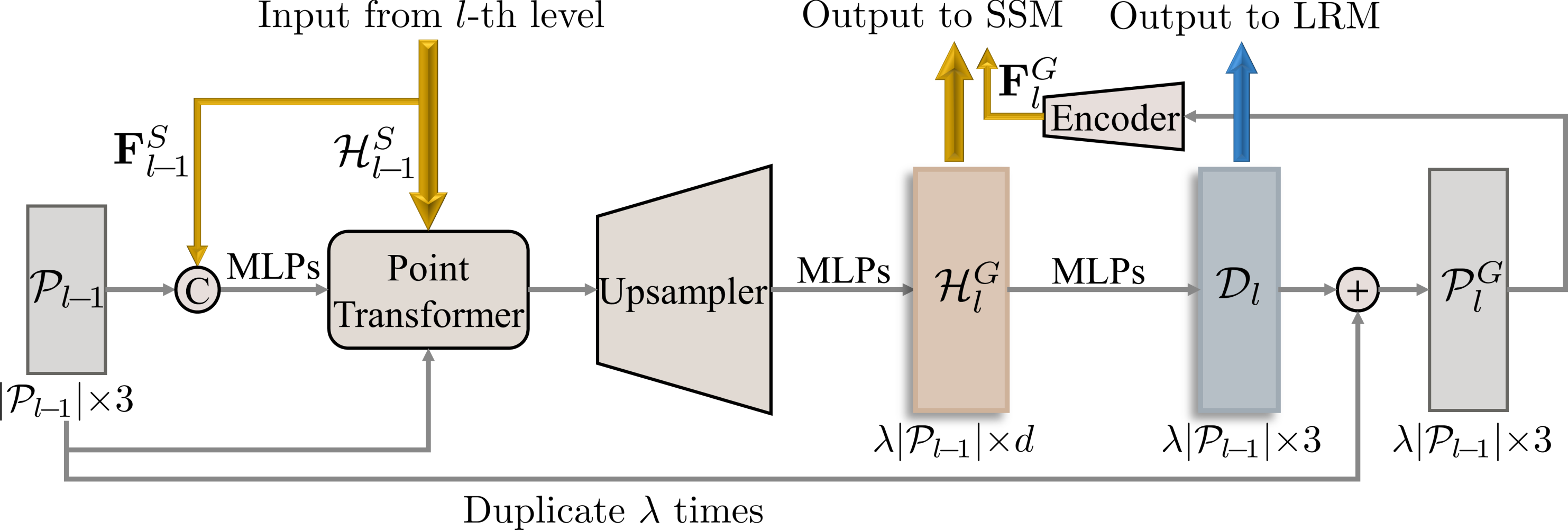}
	\caption{The detailed architecture of global completion module (GCM). Given $\mathcal{P}_{l-1}$ from previous level, GCM generates the completed scene $\mathcal{P}_l^G$ via upsampling and displacement regression. Note that, when $l=1$, the inputs $\mathbf{F}_{l-1}^{SE}$ and $\mathcal{H}_{l-1}^S$ are replaced by $\mathbf{F}$ of level \#0.
	}
	\label{fig:GCM}
\end{figure}

The purpose of global completion module (GCM) is to produce a completed point cloud $\mathcal{P}_l^{G}$ with the upsampling ratio of $\lambda$ in the $l$-th level, given the previous level's output $\mathcal{P}_{l-1}$.
Figure~\ref{fig:GCM} shows the detailed architecture, which is inspired by the snowflake point deconvolution \cite{xiang2022snowflake}.
%

Specifically, when $l\geq2$, besides using $\mathcal{P}_{l-1}$ as the input in the $l$-th level, GCM also consumes $\mathbf{F}_{l-1}^{SE}$ and $\mathcal{H}_{l-1}^S$ as inputs, which encodes the scene-wise features and the semantic features of previous level, respectively.
Then, we employ a point transformer~\cite{zhao2021point} to fuse the three kinds of inputs together.
In this way, the obtained features encode not only the input point features (brought by $\mathcal{P}_{l-1}$), but also the rich spatial and semantic information of previous level (brought by $\mathbf{F}_{l-1}^{SE}$ and $\mathcal{H}_{l-1}^S$).
Note that, when $l$$=$$1$, there is no $\mathbf{F}_0^{SE}$ and $\mathcal{H}_0^S$; see Figure~\ref{fig:architecture} as an illustration.
Hence, in the first level, besides $\mathcal{P}_0$, we further input $\mathbf{F}$ from level \#0 to GCM.
Next, an upsampler~\cite{xiang2022snowflake} is employed to upsample the fused features with $|\mathcal{P}_{l-1}|$ elements to get features $\mathcal{H}_l^G \in \mathbb{R}^{\lambda|\mathcal{P}_{l-1}| \times d}$, where $|\cdot|$ denotes the number of points and $d$ is the number of feature channels.
Considering the fact that it is difficult for the network to directly predict the absolute point coordinates with diverse and wide distribution in 3D space, 
we here choose to regress per-point displacements $\mathcal{D}_l$ from $\mathcal{H}_l^G$, which are then added to the duplicated input points to obtain the final output $\mathcal{P}_l^G$. 

Note that, to facilitate accurate semantic segmentation in each level, we further employ an encoder to extract features $\mathbf{F}_l^G$ from $\mathcal{P}_l^G$.
We design this encoder to be the same as the encoder in level \#0.
Both $\mathcal{H}_l^G$ and $\mathbf{F}_l^G$ will be fed into this level's SSM, and $\mathcal{D}_l$ will be fed into this level's LRM.

\subsection{Semantic Segmentation Module}

The purpose of semantic segmentation module (SSM) is to predict the per-point semantic labels $\mathcal{L}_{l}^S$ for $\mathcal{P}_{l}^G$.
Recall that, we generate $\mathcal{P}_{l}^G$ in GCM by adding the displacement vectors $\mathcal{D}_l$ to the duplicated $\mathcal{P}_{l-1}$.
A common case is that a point may be moved from one object to another, while the two objects belong to different classes.
In this case, if we still follow the common routine to treat each point label as a scalar value, the ground-truth semantic labels associated with some points may produce abrupt situations that cause instability in network training.
To avoid this case, instead of encouraging our network to predict a scalar semantic value for each point, we propose to regress the probability of each class.
Specifically, $\mathcal{L}_{l}^S=\{L_l^S(i)=[\mathbf{p}_{i,1}, \cdots, \mathbf{p}_{i,c}, \cdots, \mathbf{p}_{i,C}]\}$, where $L_l^S(i)$ denotes the semantic label of the $i$-th point in the $l$-th level, and $\mathbf{p}_{i,c} \in \mathbb{R}$ represents the probability of the $i$-th point belonging to the $c$-th semantic class.

\begin{figure}[t]
	\centering
	\includegraphics[width=\linewidth,scale=1.00]{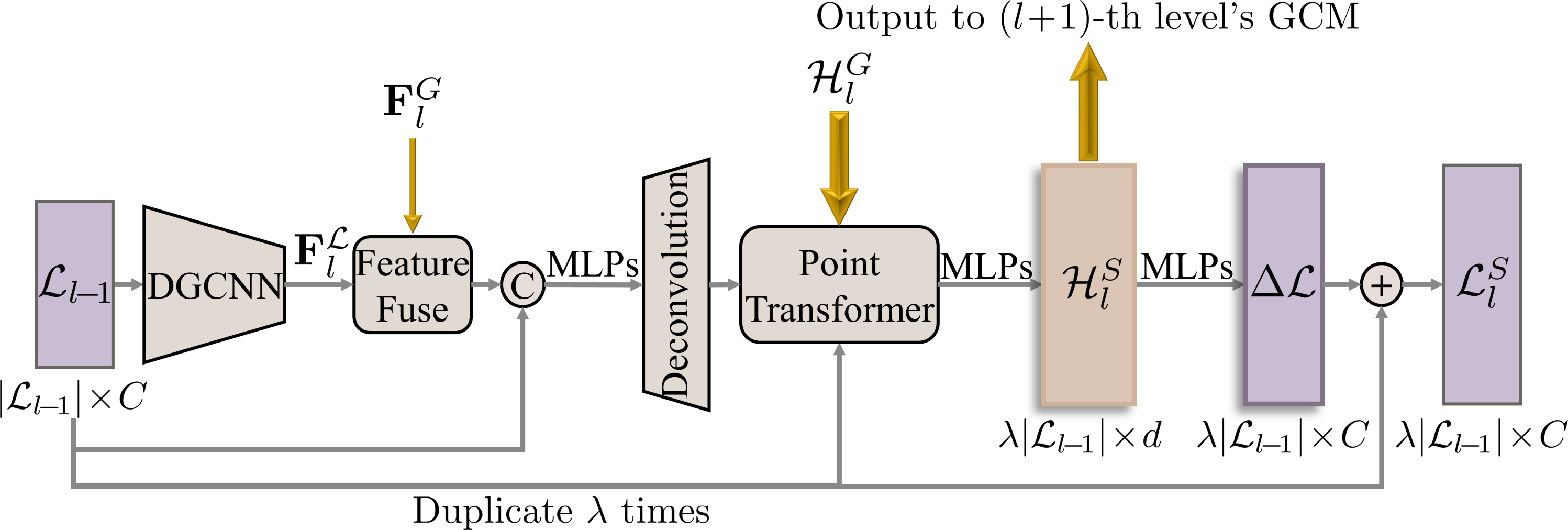}
	\caption{The detailed architecture of semantic segmentation module (SSM). Given $\mathcal{L}_{l-1}$ from previous level, SSM outputs the per-point labels for current level's completed point $\mathcal{P}_{l}^G$ by predicting label transition probability.}
	\label{fig:SSM}
\end{figure}

Figure~\ref{fig:SSM} shows the detailed architecture of our designed SSM.
Given the semantic labels (probability) $\mathcal{L}_{l-1}$ from previous level, we first extract features $\mathbf{F}_l^{\mathcal{L}}$ by using DGCNN~\cite{wang2019dynamic}.
Intuitively, $\mathbf{F}_l^{\mathcal{L}}$ encodes the probability distribution of semantics.
Next, we fuse $\mathbf{F}_l^{\mathcal{L}}$ and the geometry features $\mathbf{F}_l^G$ produced by GCM together via attention-based feature fusion~\cite{vaswani2017attention}. 
The fused features provide global scene information, which can be used to obtain coarse per-point class probability by concatenating with $\mathcal{L}_{l-1}$ and passing through MLPs.
After that, we employ deconvolution to upsample the coarse features with ratio $\lambda$. 
Considering that the label prediction highly relates to the point moving process during completion, we thus inject $\mathcal{H}_l^G$ which contains the hidden information of $\mathcal{D}_l$ into a point transformer to implicitly introduce the displacement information of this level's GCM to SSM.
Further, the transformer also takes the duplicated $\mathcal{L}_{l-1}$ as an extra input to generate the upsampled per-point class probability features $\mathcal{H}_l^S \in \mathbb{R}^{\lambda |\mathcal{L}_{l-1}| \times d}$.
%
With $\mathcal{H}_l^S$ in hand, traditional semantic segmentation models directly predict per-point labels. However, this manner ignores the relationship between point displacements and label changes.
Thus, similar to GCM,
we regress the label transition probability $\Delta \mathcal{L} \in \mathbb{R}^{\lambda|\mathcal{L}_{l-1}| \times C}$ from $\mathcal{H}_l^S$ and obtain the per-point labels by adding $\Delta \mathcal{L}$ to the duplicated $\mathcal{L}_{l-1}$.
In addition, the hidden layer output $\mathcal{H}_l^S$ will be further fused into next level's GCM for finer points.

\subsection{Local Refinement Module}

\begin{figure}[t]
	\centering
	\includegraphics[width=\linewidth,scale=1.00]{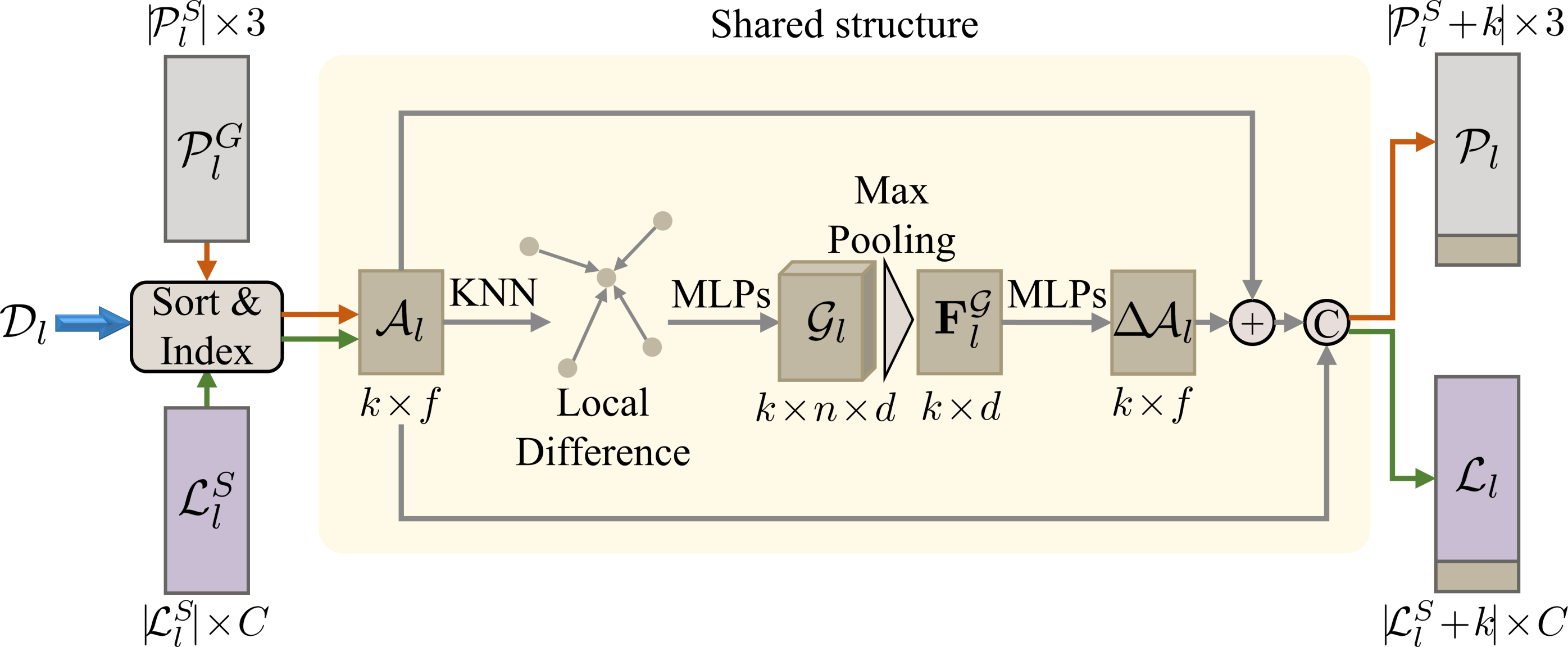}
	\caption{The detailed architecture of local refinement module (LRM). Given $\mathcal{P}_l^G$ or its associated labels $\mathcal{L}_l^S$, LRM regards the regions where points have large displacements as missing regions, then predicts minor offsets $\Delta \mathcal{A}_l$ on these points to refine the missing local regions again, and finally outputs the refined results $\mathcal{P}_l$ or $\mathcal{L}_l$.}
\label{fig:LRM}
\end{figure}

The purpose of local refinement module (LRM) is to further refine the coarse completed points $\mathcal{P}_l^G$ and its associated labels $\mathcal{L}_l^S$ from a local perspective.
As shown in Figure~\ref{fig:LRM}, LRM takes $\mathcal{P}_l^G$ or its labels $\mathcal{L}_l^S$ as well as the displacement vectors $\mathcal{D}_l$ as inputs and outputs the refined points $\mathcal{P}_{l}$ or refined labels $\mathcal{L}_{l}$. 
%
The key idea behind LRM comes from the assumption that a large displacement means that its corresponding point is most likely moved to the missing area. 
Thus, $\mathcal{D}_l$ can supply inductive knowledge and help locate large missing areas, which is also verified in our Experiment section.
Hence, we design LRM to predict minor offsets again to tune the coarse completed points and their labels in the located missing regions, thereby optimizing the associated local structures and labels with finer details.

Let's take $\mathcal{P}_l^G$ as an example to explain the local refinement process, and the process of refining $\mathcal{L}_l^S$ is the same with shared structure. 
Specifically, as shown in Figure~\ref{fig:LRM}, we first compute and sort the $L2$-norm of every displacement vector in $\mathcal{D}_l$. 
Then, we index $\mathcal{P}_l^G$ corresponding to the top $k$ biggest displacement vectors. 
We here use agents $\mathcal{A}_l \in \mathbb{R}^{k\times f}$ to denote the selected points from $\mathcal{P}_l^G$, thus $f=3$.
Note that, $f$$=$$C$ when refining labels.
To get the local structure around each reference point in $\mathcal{A}_l$, we employ KNN to group its $n$ nearest points in $\mathcal{P}_l^G$.
Next, we follow PointNet++~\cite{qi2017pointnet++} to translate the coordinates of points in a local region into a local frame relative to the reference point, and further employ MLPs to extract per-point features in each group, which is denoted as $\mathcal{G}_l \in \mathbb{R}^{k\times n\times d}$. 
Then we employ max pooling to obtain the local structure features $\mathbf{F}_l^{\mathcal{G}}$.
Similar to GCM, we also regress the offsets $\Delta \mathcal{A}_l \in \mathbb{R}^{k\times f}$ for agents, i.e. displacements of points or transition probability of labels. 
Then $\mathcal{A}_l$ is tuned by adding $\Delta \mathcal{A}_l$. 
Finally, the refined output $\mathcal{P}_{l}$ is obtained by concatenating the tuned $\mathcal{A}_l$ to the input $\mathcal{P}_l^G$. 
Naturally, the size of $\mathcal{P}_{l}$ is increased by $k$.

\subsection{Loss Function}
To fully guide network training in an end-to-end manner, we supervise both the completion and segmentation results in every level, i.e. from level \#1 to level \#3. In each level, we uniformly downsample ground truth points and their labels from the given ground truths in a dataset to the same number as each level's output, which is denoted as $\mathcal{\overline{P}}_l$ and $\mathcal{\overline{L}}_l$, respectively.
For the completion task, we employ the widely-used  Chamfer Distance (CD) to measure the difference between the completed points $\mathcal{P}_l$ and the ground truth points $\mathcal{\overline{P}}_l$:
%
%
\begin{equation}
\mathcal{L}_{CD} = \frac{1}{|\mathcal{P}_l|} \sum_{x\in\mathcal{P}_l} \mathop{\min}_{y\in\mathcal{\overline{P}}_l} \Vert x-y\Vert + \frac{1}{|\mathcal{\overline{P}}_l|} \sum_{y\in\mathcal{\overline{P}}_l} \mathop{\min}_{x\in\mathcal{P}_l} \Vert y-x\Vert,
\label{eq:CD}
\end{equation}
where $x$ and $y$ are the points in $\mathcal{P}_l$ and  $\mathcal{\overline{P}}_l$, respectively.

For the semantic segmentation task, we should note that there is no one-to-one correspondence between $\mathcal{P}_l$ and  $\mathcal{\overline{P}}_l$.
Thus, we cannot directly use the traditional cross-entropy loss as other segmentation works.
However, the lucky thing is that such one-to-one mapping has already been established when calculating Eq.~\ref{eq:CD}.
We thus leverage such correspondence to locate the nearest ground-truth point for each predicted point and use their labels for semantic loss computation.
Further, it is common that objects in a scene have various sizes, thus leading to different numbers of points on different objects.
For example, the point number in class of Floor or Wall is much larger than that in class of Table or Chair.
Hence, the semantic segmentation task often suffers from critical class imbalance. 
To relieve the overfitting to easy examples (e.g., Floor or Wall), we employ focal loss~\cite{lin2017focal} as our semantic segmentation loss:
\begin{equation}
\mathcal{L}_{sem} = \frac{1}{|\mathcal{L}_l|} \sum_{l\in\mathcal{L}_l} -(1-l)^{\gamma} \log(l),
\label{eq:focal}
\end{equation}
where $l$ denotes the predicted labels gathered by ground-truth labels $\overline{l} \in \mathcal{\overline{L}}_l$, and $\gamma$ is focusing parameter~\cite{lin2017focal}, which is increased during training to gradually increase the importance of hard examples with fewer points. 

The overall loss $\mathcal{L}_{SSC}$ consists of both $\mathcal{L}_{CD}$ and $\mathcal{L}_{sem}$, which are balanced with parameter $\alpha$:
\begin{equation}
\mathcal{L}_{SSC} = \sum_{l=1}^3 \mathcal{L}_{CD}(\mathcal{P}_l,\mathcal{\overline{P}}_l) + \alpha \mathcal{L}_{sem}(\mathcal{L}_l,\mathcal{\overline{L}}_l).
\label{eq:loss}
\end{equation}

\section{Experiments and Results}
\label{sec:experiment}


\begin{figure*}[t]
\centering
\includegraphics[width=1\textwidth]{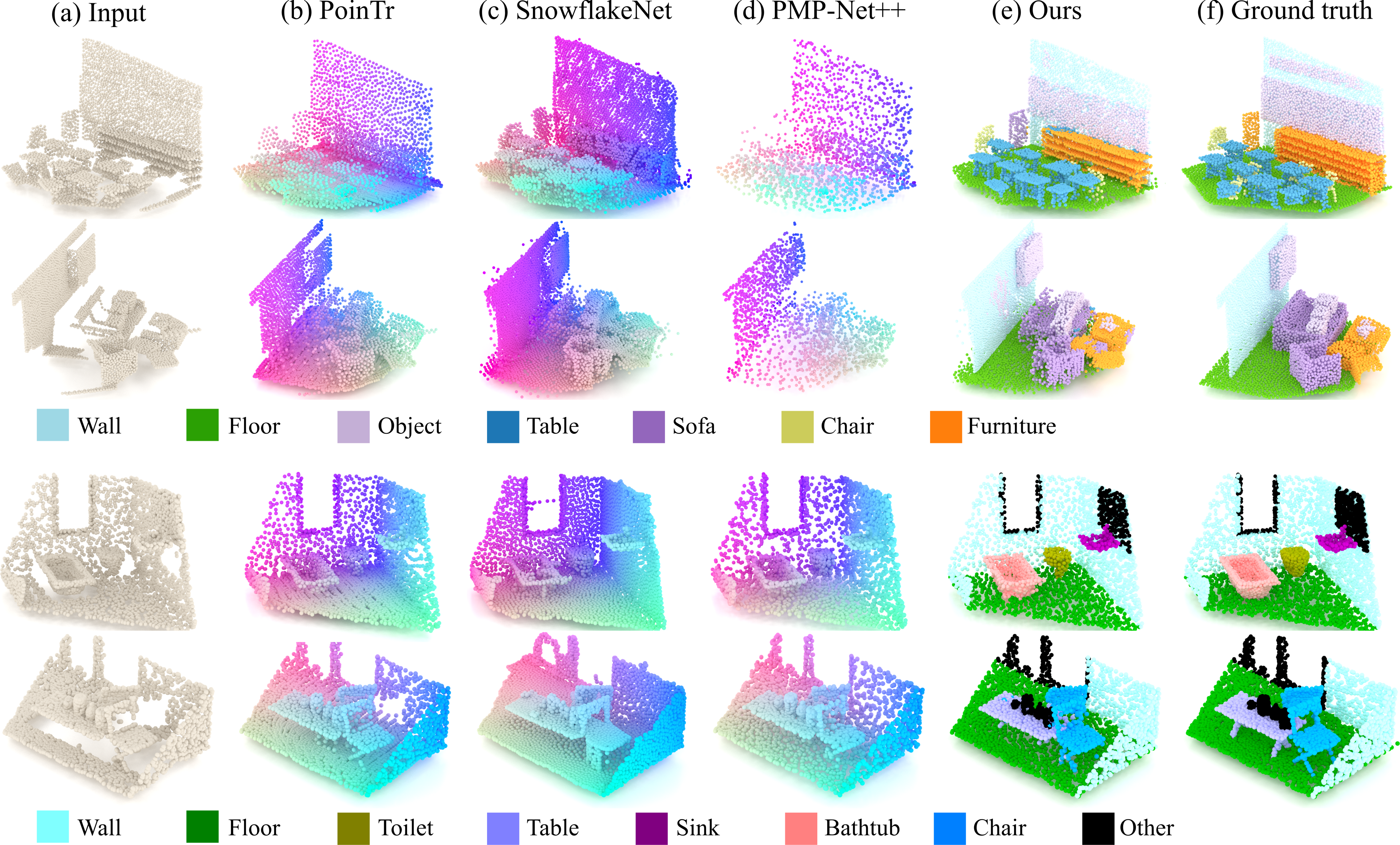}
\caption{Comparing the scene completion results of  our method (e) and recent single object point completion methods (b-d).}
\label{fig:Exp}
\end{figure*}

\subsection{Experimental Settings}

\para{Dataset preparation.} \
As there is no public dataset for point cloud semantic scene completion, we prepared two datasets, namely SSC-PC and NYUCAD-PC.
Specifically, SSC-PC is compiled based on the dataset provided in \cite{zhang2021point}.
This work provides a total of 1941 scenes covering 16 object categories in the data format of RGB-D images with associated ground-truth point clouds.
However, they do not provide camera intrinsics, so input partial point clouds cannot be obtained directly from RGB-D images.
Hence, we generate the input point clouds from ground truth points by approximating the visibility of point clouds~\cite{katz2007direct} from given views.
We follow \cite{zhang2021point} to keep 4096 points for both network input and output.
Finally, we randomly split the 1941 scenes with 1543 for training and the remaining 398 for testing.

NYUCAD-PC is derived from NYUCAD dataset~\cite{firman2016structured}, which contains 1449 RGB-D images with 795 training samples and 654 testing samples.
The partial input with 4096 points are generated from depth images by using camera intrinsics. The ground truths with 8192 points are sampled from CAD mesh annotations~\cite{guo2015predicting} with Poisson sampling algorithm.
The object categories in ground truths are mapped following~\cite{handa2016understanding}. 
Note that  the ``empty'' class is excluded, thus the number of categories is 11.
Because of severe occlusion between objects and incomplete vision of scenes, NYUCAD-PC is much more challenging than SSC-PC.

\para{Evaluation metrics.} \  
To evaluate the scene completion performance, we employ the widely-used Chamfer Distance (CD) as evaluation metric.
To evaluate the semantic segmentation performance, we employ the mean class IoU (mIoU) and the mean class accuracy (mAcc) as metrics.

\para{Implementation details.} \ 
We implement our network in PyTorch and train it on a single Nvidia RTX 3090 GPU for 400 epochs with batch size of 4.
We use Adam optimizer. The learning rate is initialized as 0.001 and decayed every two steps with 0.99 rate. 
The modulating factor $\gamma$ in Eq.~\ref{eq:focal} ranges from 0 to 5 with step size of 0.5 every 30 epochs.
The balancing parameter $\alpha$ in Eq.~\ref{eq:loss} is 0.1.
In experiments, the number of points to be refined in LRM is 96, 64 in level \#1 and \#2, respectively. The input is downsampled to 2048 in level \#0.
In NYUCAD-PC dataset, we cascade five levels and set the refined point number as 32 in LRM of level \#3.

%

\subsection{Quantitative and Qualitative Comparisons}

\para{Comparing with semantic scene completion methods.} \
We first compared our CasFusionNet with all existing point-based SSC networks, including PCSSC-Net~\cite{zhang2021point} and \cite{wang2022learning}. 
For \cite{wang2022learning}, we directly trained their released code on our prepared two datasets.
For PCSSC-Net, it demands a 12-dimension feature vector including
point coordinates, RGB values, normal vector, and
differential coordinates as network input, so their released code cannot be directly trained on our prepared datasets with only point coordinates.
However, our SSC-PC dataset is prepared from the original paper of PCSSC-Net, that shares the same 3D scenes, but with different point sampling. 
We thus directly used the evaluation values published in their original paper for comparison.
Strictly speaking, such comparison is actually unfair for our method, since the CD value calculated between our prepared input and ground truth is much larger than that in \cite{zhang2021point}, indicating that our prepared SSC-PC is more difficult.

\begin{table}[t]
\centering
\resizebox{1.0\linewidth}{!}{%
	\begin{tabular}{l|ccc|ccc}
		\toprule
		\multirow{2}*{Method} & \multicolumn{3}{c|}{SSC-PC} & \multicolumn{3}{c}{NYUCAD-PC} \\ \cmidrule{2-7}
		& CD & mIoU&mAcc&CD & mIoU&mAcc \\ 
		\midrule
		PCSSC-Net & 1.58 & 88.2 & - & - & - & -\\
		Wang et al. 2022 & 6.72& 7.4 & 10.1 &23.45 &10.2 & 14.0 \\ \midrule
		\textbf{Ours}  & 0.70& \textbf{91.3} & \textbf{94.8} & 10.28 & \textbf{49.5} & 59.7 \\
		\textbf{Ours} (without FPS)  & \textbf{0.67} & 91.2 & 94.8 & \textbf{9.99} & 49.5 & \textbf{59.8}\\
		\bottomrule
	\end{tabular}
}
\caption{Comparing semantic scene completion results of our method against recent SSC methods. Ours without FPS means we directly use the output of last level without FPS for evaluation. L1 and L2 Chamfer Distance (multiplied by $10^3$) are used in NYUCAD and SSC-PC, respectively.
}
\label{tab:SSC}
\end{table}

Table~\ref{tab:SSC} shows the comparison results. 
Clearly, our method (last two rows) achieves the best values across all metrics on both datasets.
Particularly, when we remove FPS in the last level (see Figure~\ref{fig:architecture}) and directly use the raw output for evaluation, the performance is further improved; see the bottom row.
Note that, the method \cite{wang2022learning} fails to produce a satisfying performance.
We think that this is because our prepared inputs are quite incomplete and sparse, thus preventing their proposed encoder-decoder architecture to capture a representative feature of a scene. 


\begin{table}[t]
\centering
\resizebox{0.96\linewidth}{!}{
	\begin{tabular}{l|cc}
		\toprule
		\multicolumn{1}{c|}{Method} & \multicolumn{1}{c}{SSC-PC} & \multicolumn{1}{c}{NYUCAD-PC} \\
		\midrule
		PoinTr~\cite{yu2021pointr} & 11.85 & 14.91 \\
		SnowflakeNet~\cite{xiang2022snowflake}  & 13.72& 11.83\\
		PMP-Net++~\cite{wen2022pmp} & 10.74& 13.84 \\
		\midrule
		\textbf{Ours}  & 8.96 & 10.28\\
		\textbf{Ours} (without FPS)   & \textbf{8.73}& \textbf{9.99}\\
		\bottomrule
\end{tabular}}
\caption{Comparing scene completion results with recent single object completion methods in terms of L1 Chamfer Distance (multiplied by $10^3$).}
\label{tab:SC}
\end{table}

\para{Comparing with single object completion methods.} \ 
Since most point cloud completion methods focus on a single object, we thus compared our CasFusionNet with newly proposed single object point completion methods, i.e. PoinTr~\cite{yu2021pointr}, PMP-Net++~\cite{wen2022pmp} and SnowflakeNet~\cite{xiang2022snowflake}, in terms of the scene completion performance.
In detail, we trained their networks on our prepared datasets.
Though our method further performs semantic segmentation, we only compare the CD value.
As shown in Table~\ref{tab:SC}, even with FPS, our method still achieves the lowest CD values with a significant margin compared to others, showing that directly employing single object completion methods to complete a 3D scene is not workable.

Figure~\ref{fig:Exp} further shows the qualitative comparisons on two datasets, where the top two scenes are from NYUCAD-PC and the bottom two scenes are from SSC-PC.
Clearly, regardless of the segmentation results, the completion results produced by our method (e) is the closest to the ground truths (f) against others (b-d), especially on the local details.

\subsection{Ablation Study}


To evaluate the effectiveness of the major components in
our method, we conducted an ablation study by simplifying
CasFusionNet in the following cases.
\begin{itemize}
\item \emph{Model A}: we remove both LRM and SSM  in all levels.
\item \emph{Model B}: we remove only LRM in all levels.
\item \emph{Model C}: we remove the feature fusion between GCM and SSM. More specifically, the features of previous level's GCM are fed into the next level's GCM, and the features of previous level's SSM are fed into the next level's SSM. 
\item \emph{Model D}: we remove levels \#2 \& \#3, and only keep levels \#0 \& \#1.
\end{itemize}
We re-trained the network model separately for each case using the same training dataset of NYUCAD-PC, and the evaluation results are summarized in Table~\ref{tab:ablation}.
By comparing Model A \& B vs. our full pipeline, we can see that both LRM and SSM contribute to better performance, particularly on the scene completion task.
By comparing Model C with our full pipeline and Model B, we can see that removing feature fusion results in poor performance on both scene completion and semantic segmentation.
Especially, the performance of semantic segmentation in Model C has a serious setback, thus validating the importance of our designed feature fusion. 
By comparing Model D with our full pipeline, it shows that our cascaded network design with multiple levels certainly improves task learning ability.

\subsection{Network Analysis and Discussions}
\begin{table}[t]
\centering
\resizebox{0.88\linewidth}{!}{
	\begin{tabular}{l|c|cc}
		\toprule
		\multirow{2}*{Model}& \multicolumn{1}{|c|}{Scene completion} & \multicolumn{2}{c}{Semantic segmentation} \\
		\cmidrule{2-4}
		& CD ($\times 10^{-3}$) & mIoU & mAcc \\
		\midrule
		A & 10.73 & - & -  \\
		B & 10.50 & 49.0 & 59.8 \\
		C & 10.61 & 40.3 & 51.9\\
		D & 11.60 & 41.5 & 52.2\\
		\midrule
		\textbf{Ours}  & \textbf{10.28} & \textbf{49.5} & \textbf{59.8} \\
		\bottomrule
\end{tabular}}
\caption{Ablation analysis of our network on NYUCAD-PC. 
}
\label{tab:ablation}
\end{table}
\begin{figure}[t]
\centering
\includegraphics[width=0.9\linewidth,scale=1.00]{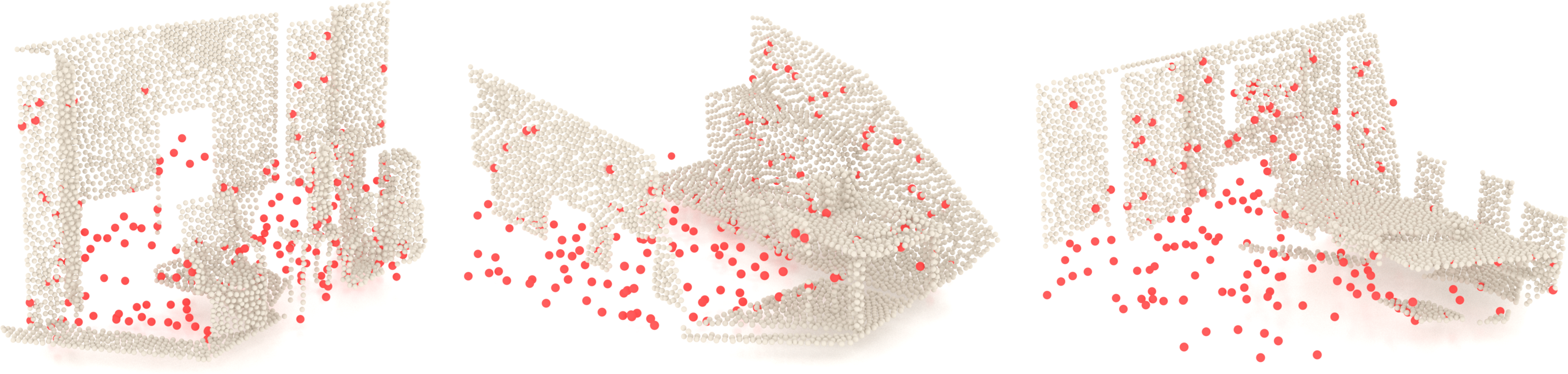}
\caption{The points with large displacement values are mostly located in missing areas; see these red points.}
\label{fig:ablation}
\end{figure}
\para{Visualization of the located missing regions.} \
As mentioned in Method section, we regard the points with large displacement values as missing regions.
To validate this, we visualize the top $k$ points back in the partial input scene; see Figure~\ref{fig:ablation} as an illustration, where the $k$ points are rendered in red color.
Clearly, most of the red points are located in missing areas, thus validating the rationality of our assumption.

\para{Limitations.} \ 
First, as a common drawback of existing works, our CasFusionNet still fails to recover local details for heavily occluded objects.
Second, our network extracts both scene-wise and point-wise features, but the object-wise features are not fully utilized, which might be useful for better completing objects. 
At last, our current network can handle about 6 scenes each with 4096 points in one second. The time performance may be further improved if we design a lightweight feature extractor in the future.

\section{Conclusion}
\label{sec:conclusion}

In this work, we present CasFusionNet, a novel point cloud semantic scene completion network by dense feature fusion.
By cascaded fusing the geometry and semantic information, our network jointly completes the missing areas and predicts the per-point semantic labels of scenes. 
Considering that there is no public dataset for point-based SSC, we thus prepared two datasets.
Both quantitative and qualitative results show that our network outperforms state-of-the-arts significantly. 
In the future, we shall explore the possibility of utilizing object-level features based on the predicted semantic labels for precise object completion. 


\section{Acknowledgments}
This work is supported by the China National Natural Science Foundation No. 62202182, No. 62276109, No. 62176101.

\bibliography{aaai23}

\end{document}